# Adaptive Extreme Learning Machine for Recurrent Beta-basis Function Neural Network Training

Naima Chouikhi, *Member, IEEE,* Adel M. Alimi, *Senior Member, IEEE*

*Abstract*— Beta Basis Function Neural Network (BBFNN) is a special kind of kernel basis neural networks. It is a feedforward network typified by the use of beta function as a hidden activation function. Beta is a flexible transfer function representing richer forms than the common existing functions. As in every network, the architecture setting as well as the learning method are two main gauntlets faced by BBFNN. In this paper, new architecture and training algorithm are proposed for the BBFNN. An Extreme Learning Machine (ELM) is used as a training approach of BBFNN with the aim of quickening the training process. The peculiarity of ELM is permitting a certain decrement of the computing time and complexity regarding the already used BBFNN learning algorithms such as back-propagation, OLS, etc. For the architectural design, a recurrent structure is added to the common BBFNN architecture in order to make it more able to deal with complex, non-linear and time varying problems. Throughout this paper, the conceived recurrent ELM-trained BBFNN is tested on a number of tasks related to time series prediction, classification and regression. Experimental results show noticeable achievements of the proposed network compared to common feed-forward and recurrent networks trained by ELM and using hyperbolic tangent as activation function. These achievements are in terms of accuracy or robustness against data breakdowns such as noise signals.

## I. Introduction

THE appeal to machine learning is resurged owing to reasons related to the high popularity of data mining and analysis. In fact, in a world full of available data varieties, computational processing seems to be very useful as it is cheap and powerful and it ensures affordable data handling [1] [2]. The automatic data treatment has provided quick and accurate models which are capable to manipulate much more complex data then deliver more precise results.

They perform not only on small data but also on very large scale ones [3]. Artificial neural networks (ANNs) have been ceaselessly exploited in several machine learning fields such as regression, classification, forecasting [4] [5] [6], etc. ANNs are characterized by a simple yet powerful generalization power.

All the authors are with REGIM-Lab: REsearch Groups in Intelligent Ma- chines, University of Sfax, National Engineering School of Sfax (ENIS), BP 1173, Sfax, 3038, Tunisia.
Email: {naima.chouikhi, adel.alimi}@ieee.org

Thus, a big attention from both of past and current researchers has been drawn around them. At the beginning, Rosenblatt designed the ancestor of artificial neuron known by the perceptron [7]. This last presented an intelligent tool for supervised data classification.

However, the perceptron was not able to perform well in many complex non-linear tasks. Thus, other neuronal models and algorithms have been developed to deal with these tasks. As for instance, back-propagation algorithm has been one of the most popular training methods [8].

Kernel Basis Function Neural Networks (KBFNNs) are among the most distinguished kinds of ANNs [9]. In spite of their structure simplicity, KBFNNs have been a good alternative to multi-layer perceptron in numerous tasks. For example, they have proven a marked performance in classification and regression tasks. All depends on the training paradigm. This last includes good kernel miscellany in order to get sophisticated hidden activations and suitable output weights estimation. The problem of selecting hidden neurons and estimating their parameters has been studied for a number of works.

Radial Basis Function Neural Networks (RBFNNs) are one of the most attractive KBFNNs that have been catching many researches. They are characterized by a sigmoid kernel. A considerable number of studies has been conducted to provide a well setting of RBFNN such as determining the network size and the centers of the sigmoid kernel function [10] [11].

Overall, Gaussian, multi-quadric, thin plate and inverse multi-quadric functions [12] are among the commonly used kernels. There is not any rule in picking up the most suitable kernel. It is up to the problem to be handled. For example, a radial basis function is known to be mainly problem specific. In conventional RBFNN, the Euclidean distance between the input feature and neurons' center vectors is used as evaluation criterion [10]. However, it may occur that this distance is not the most dominant measure that separates the different classes.

Beta Basis Function Neural Networks (BBFNNs) [13] [14] belong to the KBF Neural Networks as well as RBF Neural Networks. BBFNN is a well-known neural architecture which was initially proposed by Alimi in 1997 [15]. Since its design, BBFNN has realized numerous successes in dealing with many tasks such in [16] [17] [18]. It has created a favorable alternative for function approximation and many other problems.

BBFNN is three layered. A linear function is usually applied at the level of output neurons. The specificity of such a network resides in the transfer function as it uses a special kind of activation function called "Beta function" [15] [19]. This last has been used as an activation function thanks to several virtues. In fact, it presents a high flexibility and a universal approximation capability. Moreover, it generates richer shapes (asymmetry, linearity, etc). According to a well determined parameters setting, Beta function can approximate sigmoid function. Since its design, the BBFNN has been of wide use.

BBFNN contains three layers including the input, the hidden BBF and the output layers. The inputs, after being multiplied by a set of weights, go through beta based activation at the level of the hidden layer where each hidden neuron is characterized by a set of parameters [20]. The final outputs are obtained after a linear combination of the hidden layer activations. As in every network, the challenges faced by BBFNN are related to both of the training method and the architecture fixing.

Researchers focusing in BBFNN training have supported a number of algorithms. The orthogonal least squares (OLS) algorithm [21] has been used for the training of KBFNN generally and BBFNN especially [22]. It is an iterative algorithm where just one center is concerned at each step. OLS is an elegant algorithm assuring not only the parametric setting but also the structural configuration of the neural network. Nevertheless, this method has several limits. Indeed, the manner by which the parameters of the candidate beta activation functions centers are picked up is not strong [22]. Usually, the center vectors are taken from a number of input vectors already existing in the training set. Dhahri *et al.* utilized the famous Back-Propagation algorithm (BP) to train the BBFNN. The idea consists to stare the number of hidden neurons and to settle the beta parameters based on the gradient descent approach. The used training method permits the network to grow-up by squirting new neurons progressively. Many variants of evolutionary algorithms have been also used to train BBFNN [20] [22] [17] [19]. Although the already cited methods have given competitive results, they result on a complexity increase. The iterative computation provided by these methods make them greedy in terms of runtime.

Extreme learning machine (ELM) [8] is a non-iterative training method. It is among the least squares-based methods that are conceived to train feed-forward neural network [23]. Since it was proposed by Huang [10] in 2006, ELM has acquired a huge popularity. ELM has realized successive successes in training ANNs [24]. Kernel learning has been granted to ELM to procure enhanced generalization capability while keeping the user intervention limited. The input weights are randomly generated and fixed throughout the learning process. The output weights are analytically computed. Many studies have shown that the ELM achieves better performance if well extracted hidden layer features are given [25] [26]. Thanks to its simple but efficient learning strategy, ELM is adopted, in this paper, as a training procedure of BBFNN. On the other side, recurrent neural networks (RNNs) have proved more efficiency in dealing with complex tasks compared to the feed-forward networks. RNNs enable signals from circulating in both directions (from left to right and vice versa) by conceiving loops within the network. Recurrent neural networks (RNNs) [27] [28] are dynamic and powerful even though they can get some complex computations. The connection loops provide to RNNs a kind of memory which enables them to get more information about the network past dynamics.

In this paper, a new variant of BBFNN is proposed according to the two challenges already mentioned. In fact, ELM is adopted as a training algorithm. The random transformation of the input space generated by the random input weight distribution strengthens the BBFNN inner dynamics. In order to make it more efficient in dealing with more complex problems, a set of recurrent connections is added at the level of the BBF layer. The added recurrence enables the BBFNN of performing in both temporal and non-temporal tasks.

The outline of the paper is divided into four sections. In section 2, the BBFNN is introduced. A thorough description of the beta function is done as well. In section 3, the proposed recurrent BBFNN trained by ELM is introduced. In section 4, the already defined approach is applied to various tasks including classification, time series prediction and regression. Tests are performed on a number of well-known datasets to which some breakdowns are injected. Several simulation results prove the efficiency of the new proposed network. By the end of the paper, a conclusion and outlook for future work are given.

## II. BETA BASIS FUNCTION NEURAL NETWORK (BBFNN)

An artificial neural network is presented as a set of connected neurons where each neuron performs a given function and each connection specifies the direction of passage of the signal from one neuron to another. The behavior of a neural network is governed by a number of parameters that can be modified and distributed over neurons. Each neuron has its own transfer function.

Beta Basis Function Neural Network (BBFNN) is a special kind of feed-forward networks. It is a three-layered net- work: input, output and hidden layers. The hidden layer includes N neurons undergoing a non-linear activation function [20]. This last is performed by a beta basis function. This last has been used by Alimi for the first time as a neural activation function in 1997 [15]. The architecture of a BBFNN is shown in fig. 1.

The last layer includes neurons allowing a linear activation of the hidden layer outputs weighted by output weights. The BBFNN's output is computed in equation (1).

$$y = \sum_{j=1}^{N} W^{out} \beta_j(u) \qquad (1)$$

where $y$ is the network output. $u$ is the input vector. The input weight matrix is taken to be equal to ones matrix for BBFNN. $W^{out}$ designates the output weight matrix. K, N and L are the size of the input, hidden and output layers. $\beta$ is the hidden activation function.

The Beta function's name comes from the Beta integral. Thanks to its ability to generate rich and flexible shapes (asymmetry, linearity, etc.), Beta has been used as a neural transfer function. Fig. 2 shows examples of shapes that can be generated by Beta function. By varying its parameters,

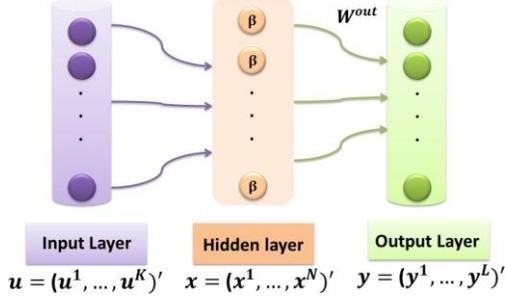

$$u = (u^1, ..., u^K)' \qquad x = (x^1, ..., x^N)' \qquad y = (y^1, ..., y^L)'$$

Fig. 1. Beta Basis Function Neural Network model

Beta has been adapted for different data distributions. Beta function in one-dimensional case is defined by equations (2), (3), (4) and (5) [22].

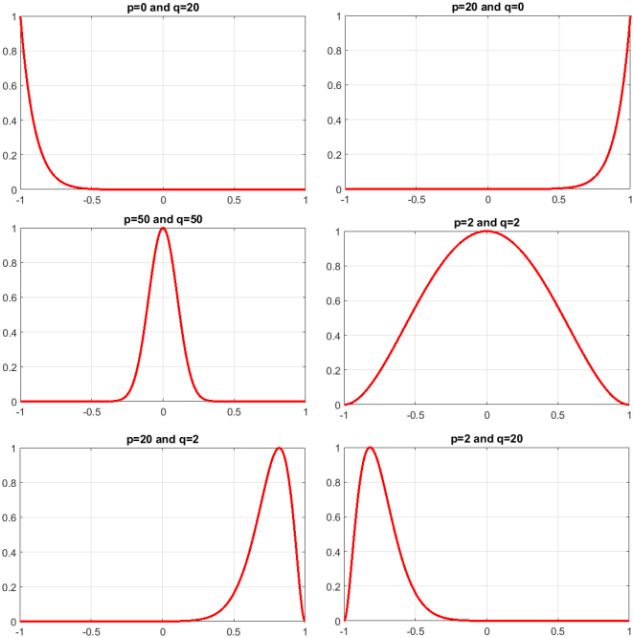

Fig. 2. Examples of Beta basis function shapes

In this case, suppose $u_0$ and $u_1$ belong to $\mathbb{R}$ such that $u_0 < u_1$. The Beta function is defined by $\beta(u) = \beta(u, u_0, u_1, p, q)$ according to four cases.

- **1st case**: $p > 0$ and $q > 0$

$$\beta(u) = \begin{cases} [\dfrac{u - u_0}{u_c - u_0}]^p [\dfrac{u_1 - u}{u_c - u_1}]^q & if\ u_0 < u < u_1 \\ 0 & else \end{cases} \quad (2)$$

where $u_c = \dfrac{pu_1 + qu_0}{p + q}$ is the beta center.

- **2nd case**: $p > 0$ and $q = 0$

$$\beta(u) = \begin{cases} [\dfrac{u - u_0}{u_c - u_0}]^p & if\ u_0 < u < u_1 \\ 0 & if\ u < u_0 \\ 1 & if\ u > u_1 \end{cases} \quad (3)$$

- **3rd case**: $p = 0$ and $q > 0$

$$\beta(u) = \begin{cases} [\dfrac{u - u_1}{u_c - u_1}]^q & if\ u_0 < u < u_1 \\ 1 & if\ u < u_0 \\ 0 & if\ u > u_1 \end{cases} \quad (4)$$

- **4th case**: $p = 0$ and $q = 0$

$$\beta(u) = 1, \forall u \in \mathbb{R} \quad (5)$$

The shape and hence behavior of the Beta-SVM function is influenced by the choice of the parameters $p$, $q$, $u_0$ and $u_1$. For example, when p = q the function is symmetric and very closely resembles a Gaussian kernel. When either $p = 0$ or $q = 0$ (but not both) its shape looks like that of a sigmoid. The variance of the function is also affected by the choice of $p$ and $q$: the smaller the values of $p$ and $q$ the larger the variance. Beta basis function plays the role of a linear function of $u$, if ($p = 1; q = 0$) or ($p = 0; q = 1$).

In the multi-dimensional case (dimension = K) and if $U = (u^1, ..., u^K)$, $U_0 = (u_0^1, ..., u_0^K)$, $U_1 = (u_1^1, ..., u_1^K)$, $P = (p^1, ..., p^K)$ and $Q = (q^1, ..., q^K)$ then the Beta function is defined in equation (6).

$$\beta(U, U_0, U_1, P, Q) = \prod_{i=1}^{K} \beta(u^i, u_0^i, u_1^i, p^i, q^i) \quad (6)$$

where $\beta(u^i, u_0^i, u_1^i, p^i, q^i)$ is the Beta function in the one dimensional case. The good or bad performance of the BBFNN refers to the good or bad choice of several factors such as the network architecture and the training method.

In fact, the setting of the number of hidden neurons, the beta parameters and the way to compute the output weights are very crucial to the efficiency of the network. Different are the learning methods used to deal with these issues. They are classified into three classes which are the supervised, semi-supervised and unsupervised learning strategies [29]. These parameters can be randomly or methodologically set. The setting of a suitable learning algorithm is among the main challenges in every neural network. The dare behind our model is to provide a powerful architecture trained by a fast and non-complex training algorithm for the BBFNN.

## III. RECURRENT ELM-BASED BETA BASIS FUNCTION NEURAL NETWORK (REC-ELM-BBFNN)

In this section, the proposed recurrent architecture and the training method of the BBFNN are presented.

### A. Extreme Learning Machine for BBFNN training

Extreme learning machines (ELM) has been proposed for generalized single hidden layer feed forward neural network

(SLFNs) [11,26,27]. Unlike other neural networks with back-propagation (BP) [28], the hidden nodes in ELM are randomly generated, as long as the activation functions of the neurons are non-linear piecewise continuous. In fact, it is characterized by a simple training process during which the synaptic weights from the input to the hidden layer are randomly set then still fixed. Only the output weights are trained generally according to an analytical approach.

This approach consists of computing the pseudo-inverse of the hidden weight matrix. This fact makes ELM quicker than several gradient based methods. Let $y$ be the network output.

$$y = \sum_{j=1}^{N} h(W^{out} x^j) \qquad (7)$$

where $x^j$ is the activation of the $j^{th}$ hidden neuron. $W^{out}$ is an output weight matrix. $h$ denotes the activation function of the output layer. It is generally a linear identity function($h(x^j) = x^j$). In real applications, $h$ can be described as in(8)

$$h(x) = x = g(W^{in}u) \qquad (8)$$

where g is an activation function satisfying ELM universal approximation capability. g is chosen as a non-linear function (generally hyperbolic tangent "tanh"). u designates the input vector and $W^{in} \epsilon \mathbb{R}^{N*K}$ is a random input weight matrix. Kernel learning was integrated into ELM to obtain better generalization with less user intervention. Let Data $= \{(U_z, Yd_z) | z = 1..M\}$ be a set of M training samples, where $U_z = (u_z^1, \dots, u_z^K)$ is a K-dimensional input vector and its respective desired output $Yd_z = (yd_z^1, \dots, yd_z^L)$ is a desired outputs vector.

The procedure of getting x is called ELM feature mapping which maps the input data from the input space $\mathbb{R}^K$ to the feature space $\mathbb{R}^N$.

$W^{out} \epsilon \mathbb{R}^{L*N}$ denotes the output weights between the hidden layer ($N$ neurons) and the output layer ($L$ neurons). It can be computed based on equation (9).

$$W^{out} = pinv(H) * Yd \qquad (9)$$

where $H \epsilon \mathbb{R}^{N*M}$ is the hidden layer output matrix obtained after computing the activation of the hidden neurons for all the training data patterns. $pinv(H)$ denotes the Moore-Penrose pseudo-inverse of the hidden activation matrix H.

The new proposed training algorithm of BBFNN in this work is ELM which added a random projection of the inputs in the feature beta-based space. In the previous BBFNN architectures, the input weights are set to ones matrix. In ELM, the parameters are randomly generated based on a continuous probability distribution. Giving different input weights values permits to give to each input a specific contribution in the next layer. In this part, the learning strategy of the BBFNN is altered.

In basic ELM which uses "tanh" as activation function, the inputs are weighted by the input weight matrix then summed together then the activation function is applied to the whole sum. If this work, as the transfer function to be applied is beta, the computation of the hidden activations is done differently (equation 10). Each input is weighed by the appropriate weight then a beta activation is applied. Thereafter, a product of the computed Beta activations is performed to obtain the activation $x^j$ of the $j^{th}$ neuron.

$$x^j = \beta(W^{in}u) = \prod_{i=1}^{K} \beta(W^{in}(j,i)u^i(t+1), u_0^i, u_1^i, p^i, q^i) \qquad (10)$$

Where $W^{in}(j,i)$ is the input synaptic weight from the $i^{th}$ unit to the $j^{th}$ hidden one. Once the hidden activations are obtained, the network output as well as the output weights can be derived by applying equations (7) and (9) respectively. In the next part, the new recurrent BBFNN architecture is described. Let it be called 'Rec-ELM-BBFNN'.

### B. Adaptive ELM-based training of recurrent BBFNN

The added value of RNNs against FNNs is the dynamism reflected by the addition of further synaptic connections. Indeed, in RNNs, all or some previous neurons' outputs become inputs in the next time step. Saving the output of a layer and feeding it back to another layer or to the same layer itself is the principle of RNN process. This fact makes RNNs having a kind of virtual memory that records previous network dynamics. This memory is created through backward synaptic loops. RNNs are then capable to deal efficiently with larger applications scale than FNNs.

Thus, a new recurrent architecture is proposed in this study for the BBFNN in order to enhance its performance in a number of tasks. In addition to the input and output weight matrices ($W^{in}$ and $W^{out}$), another recurrent weight matrix $W^{rec} \epsilon \mathbb{R}^{N*N}$ is added at the level of the hidden layer. $W^{rec}$ contains the recurrent connections that relate the hidden neurons to themselves. These newly added weights create an internal memory within the beta units and make the network capable to handle more complex non-linear tasks. Fig. 3 presents the architecture of a recurrent BBFNN.

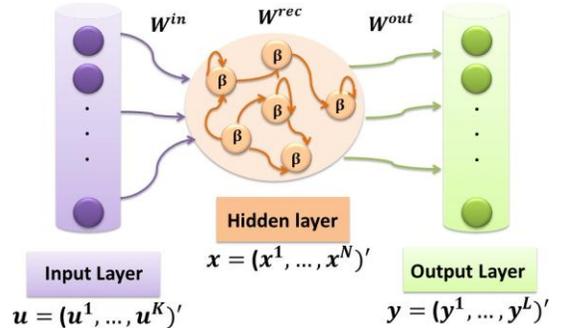

Fig. 3. Recurrent Beta Basis Function Neural Network model

Let $t$ designates the index of the dataset pattern. The new recurrent Beta layer's input inp (equation (11)), in this case,is composed of inputs from the dataset and the previous hidden states. The computation of the $j^{th}$ hidden neuron input state $x^j(t+1)$ is performed as in equation (12).

$$inp(t+1) = [u(t+1): x(t)] \qquad (11)$$

➔ $inp(t+1) = [u^1(t+1), \dots, u^K(t+1), x^1(t), \dots, x^N(t)]$

where [. : .] stands for a vertical vector concatenation.

$$x^j(t+1) = \prod_{k=1}^{K} \beta(W^{in}(j,i)u^i(t+1), u_0^i, u_1^i, p^i, q^i) *$$
$$\prod_{k=1}^{N} \beta(W^{rec}(j,k)x^k(t), u_0^k, u_1^k, p^k, q^k) \qquad (12)$$

Where $u^i(t+1)$ is the value of the ith input neuron in the $(t+1)^{th}$ data pattern. $x^k(t)$ is the previous activation of the $k^{th}$ hidden neuron. $W^{in}(j,k)$ is the synaptic weight from hidden neuron k to hidden neuron j. Hence, the network output can be computed as in equation (7).

---

**Algorithm 1: Rec-ELM-BBFNN algorithm**

**1:** Initialize the network's parameters: the hidden neurons $N$, the input and recurrent weights $W^{in}$ and $W^{rec}$
**2:** Initialize the beta units' parameters $P, Q, U_0$ and $U_1$
**3:** Initialize the first hidden states of the beta units $x(t=0)$
**For** each sample $(t+1)$ in the training dataset do
**For** each beta unit $j$ in the hidden layer do
**4:** Compute the hidden activation $x^j(t+1)$ of the corresponding $j^{th}$ beta unit based on equation (12).
**End for**
**End for**
**5:** Extract the hidden layer output matrix $H$.
**6:** Compute the output weight matrix $W^{out}$ according to equation (9).
**7:** Compute the network output according to equation (7)
**8:** Compute the training accuracy.
**9:** Compute the testing accuracy after squirting the testing Dataset.

---

dynamics and reduced runtime within the proposed model, it is recommended to choose a sparse $W^{rec}$ matrix. The sparsity degree of $W^{rec}$ designates the rate of zero synaptic connections among all the possible existent connections whereas its connectivity rate presents the number of non-zero connections. Also, it is advisable to make $W^{rec}$ scaled by its spectral radius $\rho$ which, in its turn, should be lower than 1. The spectral radius $\rho$, whose expression figures in equation (13), represents the $W^{rec}$'s eigenvalue having the higher absolute value. Therefore, there will be either densely connected neurons with smaller synaptic hidden weights or sparsely connected units with higher weights values. Stability can be ensured in both of situations. Also, the asymptotic properties of the excited hidden dynamics will be related to the driving signal. Intuitively, the hidden layer will asymptotically wash out any information from initial conditions.

$$\rho = \text{Max}(eigenvalues(W^{rec})) \qquad (13)$$

## IV. EXPERIMENTAL RESULTS AND DISCUSSION

The performance of the conceived 'Rec-ELM-BBFNN' is tracked on a number of tasks. It is applied to various well-known benchmark datasets for time series prediction, regression and classification. These datasets belong to three repositories resources which are the "Times Series Data Library" [30], the "UCR Time Series Data Mining Archive" [31] and the "KEEL Archive" [32] The specifications of the network and the beta function for each dataset are indicated in table 1.

In order to substantiate the effectiveness of our model, the following ELM-based models are implemented and evaluated too.

- **Tanh-ELM:** A feed-forward Neural Network trained by ELM with tanh as transfer function.
- **Rec-Tanh-ELM:** A recurrent feed-forward Neural Network trained by ELM with tanh as transfer function.
- **ELM-BBFNN:** A Beta basis function neural network trained by ELM.
- **Rec-ELM-BBFNN:** A recurrent Beta basis function neural network trained by ELM.

Our empirical evaluation focuses on the following couple of questions.

**Question 1.** *Efficacy:* Does the generalization capability of the proposed model, Rec-ELM-BBFNN, able to compete other intelligent models?

**Question 2.** *Robustness:* Does Rec-ELM-BBFNN achieve a competitive stable behavior in presence of breakdowns such as noise?

For the classification tasks, the classification accuracy (equation (14)) is the accuracy criterion used to evaluate the approach. Whereas for the prediction and regression datasets, the involved evaluation metric is by either the Mean Square Error (MSE) or the Root Mean Square error (RMSE). Both of the errors are computed according to the difference between the network outputs and the desired ones. The expressions of MSE and RMSE are given in equations (15) and (16), respectively.

$$CA = \frac{Number\ of\ well\ classified\ patterns}{Total\ number\ of\ testing\ patterns} \qquad (14)$$

$$MSE = \frac{1}{m} \sum_{i=1}^{m} (yd^i - y^i)^2 \qquad (15)$$

$$RMSE = \sqrt{\frac{1}{m} \sum_{i=1}^{m} (yd^i - y^i)^2} \qquad (16)$$

Where m designates the number of testing patterns, yd is the desired output and y is the network output. The remains of this part are devoted to the datasets description as well as their empirical results. In order to sleek the learning process, a linear normalization is applied to all the data, except those of regression tasks, in a way to make them being within [0 1] or [-1 1].

The implemented models are executed many times. The average of the results (CA, RMSE or MSE) obtained after this set of runs is taken as a final result. Let $U_p, L_p, U_q, L_q, U_{u_0}, L_{u_0}, U_{u_1}, L_{u_1}$ be the upper and lower values of the parameters $p, q, u_0$ and $u_1$, respectively. A set of trials and tests are performed in order to set these values.

TABLE 1: Beta function parameters setting and datasets specifications.

| | Source | Dataset | $N$ | $K$ | $L$ | #C | Length | #$Train$ | #$Test$ | $U_p$ | $L_p$ | $U_q$ | $L_q$ | $L_{u_0}$ | $U_{u_0}$ | $L_{u_1}$ | $U_{u_1}$ |
|---|---|---|---|---|---|---|---|---|---|---|---|---|---|---|---|---|---|
| 1 | [32] | Pima | 50 | 8 | 1 | 2 | 768 | 615 | 153 | 2 | 10 | 1 | 10 | -5 | 5 | -5 | 5 |
| 2 | [32] | Mammographic | 30 | 5 | 1 | 2 | 830 | 664 | 166 | 1 | 20 | 1 | 20 | -10 | 10 | -10 | 10 |
| 3 | [32] | Monk-2 | 50 | 6 | 1 | 2 | 432 | 346 | 86 | 1 | 20 | 1 | 20 | -10 | 10 | -10 | 10 |
| 4 | [32] | Titanic | 20 | 3 | 1 | 2 | 2201 | 1761 | 440 | 1 | 2 | 1 | 2 | -10 | 10 | -10 | 10 |
| 5 | [32] | Australian | 45 | 14 | 1 | 2 | 690 | 552 | 138 | 1 | 2 | 1 | 2 | -10 | 10 | -10 | 10 |
| 6 | [32] | German | 70 | 20 | 1 | 2 | 1000 | 800 | 200 | 1 | 2 | 1 | 2 | -5 | 5 | -5 | 5 |
| 7 | [32] | Thyroid | 50 | 21 | 1 | 3 | 7200 | 5760 | 1440 | 1 | 2 | 1 | 2 | -10 | 10 | -10 | 10 |
| 8 | [31] | ECG200 | 150 | 96 | 1 | 2 | 200 | 100 | 100 | 1 | 5 | 1 | 5 | -1 | 1 | -1 | 1 |
| 9 | [32] | Breast cancer | 20 | 9 | 1 | 2 | 699 | 500 | 199 | 1 | 20 | 1 | 20 | -1 | 1 | -1 | 1 |
| 10 | [31] | Earthquakes | 300 | 512 | 1 | 2 | 461 | 322 | 139 | 2 | 3 | 1 | 3 | -5 | 5 | -5 | 5 |
| 11 | [30] | Passenger | 20 | 1 | 1 | - | 144 | 115 | 29 | 1 | 20 | 1 | 20 | -5 | 5 | -5 | 5 |
| 12 | [30] | Milk | 20 | 1 | 1 | - | 168 | 124 | 44 | 1 | 20 | 1 | 20 | -5 | 5 | -5 | 5 |
| 13 | [30] | Stock | 20 | 1 | 1 | - | 168 | 124 | 44 | 1 | 10 | 1 | 10 | -5 | 5 | -5 | 5 |
| 14 | [30] | Lynx | 20 | 1 | 1 | - | 99 | 79 | 20 | 1 | 10 | 1 | 10 | -5 | 5 | -5 | 5 |
| 15 | [30] | Jenkin's Box | 20 | 2 | 1 | - | 296 | 200 | 96 | 1 | 20 | 1 | 20 | -5 | 5 | -5 | 5 |
| 16 | [30] | Sunspot | 20 | 4 | 1 | - | 315 | 220 | 95 | 1 | 5 | 1 | 5 | -10 | 10 | -10 | 10 |
| 17 | [ - ] | S & P 500 | 80 | 30 | 1 | - | 970 | 600 | 370 | 1 | 10 | 1 | 10 | -10 | 10 | -10 | 10 |
| 18 | [ - ] | Magnetic levitation | 30 | 1 | 1 | - | 4001 | 2500 | 1500 | 1 | 20 | 1 | 20 | -1 | 1 | -1 | 1 |
| 19 | [32] | Ailerons | 80 | 40 | 1 | - | 13750 | 11000 | 2750 | 1 | 2 | 1 | 2 | -5 | 5 | -5 | 5 |
| 20 | [32] | House | 50 | 16 | 1 | - | 22784 | 18228 | 4556 | 1 | 2 | 1 | 2 | -5 | 5 | -5 | 5 |
| 21 | [32] | California | 50 | 8 | 1 | - | 20640 | 16512 | 4128 | 1 | 2 | 1 | 2 | -5 | 5 | -5 | 5 |

Table 1 illustrates the selected beta parameters as well as the number of input, hidden and output units $K, N, L$ for each studied dataset. For the classification datasets, the number of classes $C$ is also mentioned. Table 1 reports also the length of the whole datasets as well as the training and testing datasets sizes. The links to the datasets web sources figure in column 2. The global model's and Beta parameters are picked out by experience dependent on the task to be handled.

The performance of any intelligent system in general and our systems in special heavily depends on the quality of the training data, but also on the robustness against noise. In order to study the robustness of the proposed approach, a Gaussian noise signal is squirted randomly in both of the training and testing databases. In fact, real-world data, which constitute generally the input of data mining approaches, may be damaged by different breakdowns such as noise disturbing. Noise is considered as meaningless data disturbing any data mining analysis [33]. It is an unavoidable problem affecting the data aggregation process.

Noise comes from either implicit measurement tools' errors or random faults generated by batch process. Three noise levels are studied in this work. Each noise signal is designated by a signal to noise ratio SNR. The noise level is inversely proportional to the SNR. The lower the SNR, the higher the noise. The SNR is expressed in dB and the three levels correspond to SNR=50dB, SNR=10dB and SNR=1dB.

### A. Classification

A set of classification problems are tackled in order to analyze the performance of the proposed network. For the datasets which are taken from the KEEL archive [32], a 5-fold cross validation method was performed for overall the tasks. In fact, each data set is partitioned into five parts where four are devoted for the training and the remaining one is consecrated for the test.

Table 2 gives the results in terms of classification accuracy for the datasets 1-7. The mean CA and Standard deviation St.D are presented in table 2. The performances of other existent methods are also included in table 2. The best CA is written in bold. The symbol "-" means that the corresponding value doesn't exist in the literature study.

According to table 2, the outperformance of Rec-ELMBBFNN is obvious for overall the datasets. This fact reveals on one side the importance of beta as activation function as there is an advance compared to the tanh-based architectures.

On the other side, the recurrent connections have brought-up an added value as there is a clear overture of our architecture compared to the feed-forward one. Many already existing variants of ELM (ScELM, SSELM, LapTELM, TROP-ELM, SBELM, etc.) and SVM (LapSVM, RVM, SVM, etc.) are included in table 2. For every dataset, our proposed recurrent beta-based classifier gives more improved results. For instance, the improvement rate overtakes 5% compared to all the mentioned literature approaches on pima dataset (except for CHELM) and 1% on the australian dataset.

Regarding the tanh-based networks, their classification accuracy degrades by almost 1% compared to those achieved by our beta-based networks on the majority of the datasets (pima, mammographic, german and thyroid).

Table 3 reports the accuracy results in terms of the classification accuracy between the proposed models and other existing approaches on the ECG200 database. According to the results listed within table 3, Rec-ELM-BBFNN performs more accurately than other methods. Under the CA metric, the amelioration achieved by ELM-BBFNN over Tanh-ELM overtakes 10% for the ECG 200 dataset. The improvement realized by Rec-ELM-BBFNN over Rec-Tanh-ELM surpasses 6%.

A statistical study of the performance of the recurrent BBFNN on the earthquakes classification problem appears in table 4. As for the previous problems, a comparison with already existing models in terms of the classification accuracy is also performed within table 4. For each of Tanh-ELM, Rec-Tanh-ELM, ELM-BBFNN and Rec-ELM-BBFNN, the testing

TABLE 2: Classification accuracy-based comparison between Tanh-ELM, Rec-Tanh-ELM, ELM-BBFNN and Rec-ELM- BBFNN (top: CA, bottom: St.D) on datasets 1-7.

| Method | Pima | Monk-2 | Mammographic | Titanic | Australian | German | Thyroid |
|---|---|---|---|---|---|---|---|
| CHELM [34] | 0.7737 - | - | - | - | 0.8544 - | 0.7960 - | - |
| WPNN [35] | 0.6882 - | 0.8761 - | - | - | - | - | - |
| RBFNN [35] | 0.7514 - | 0.7567 - | - | - | - | - | - |
| PART) [36] | 0.5044 - | 0.5027 - | 0.7722 - | 0.5001 - | 0.6443 - | 0.5000 - | - |
| 1 NN [36] | 0.6513 - | 0.7419 - | 0.7550 - | 0.5227 - | 0.8228 - | 0.6275 - | - |
| 3 NN [36] | 0.6713 - | 0.9509 - | 0.8107 - | 0.5493 - | 0.8474 - | 0.6349 - | - |
| SVM [36] | 0.6837 - | 0.9611 - | 0.8078 - | 0.6824 - | 0.8045 - | 0.7056 - | - |
| C45 [36] | 0.7047 - | - | - | 0.6911 - | 0.8449 - | 0.6303 - | - |
| LDA [36] | 0.7235 - | 0.7756 - | - | 0.6996 - | 0.8649 - | 0.6438 - | - |
| ELM [37] | 0.7279 ±0.0128 | - | - | - | 0.8571 ±0.0208 | - | - |
| ScELM [37] | 0.6988 ±0.0190 | - | - | - | 0.8565 ±0.0196 | - | - |
| SVM [37] | 0.6862 ±0.0426 | - | - | - | 0.8564 ±0.0208 | - | - |
| SS-ELM [38] | 0.6189 ±0.0252 | - | - | - | 0.7081 ±0.0129 | - | - |
| LapSVM [38] | 0.7247 ±0.0250 | - | - | - | 0.6970 ±0.0258 | - | - |
| TELM [38] | 0.5936 ±0.0438 | - | - | - | 0.6709 ±0.0384 | - | - |
| LapTELM [38] | 0.6894 ±0.0483 | - | - | - | 0.7551 ±0.0355 | - | - |
| IS-SSGA [39] | 0.7220 ±0.0350 | - | 0.7999 ±0.039 | - | - | 0.7087 ±0.0369 | - |
| IS-GGA [39] | 0.7271 ±0.0454 | - | 0.7985 ±0.0409 | - | - | 0.7073 ±0.0404 | - |
| SBELM [40] | - | - | - | - | 0.6783 ±0.0012 | 0.7730 ±0.0259 | - |
| TROP-ELM [40] | - | - | - | - | 0.6742 ±0.0197 | 0.7550 ±0.0395 | - |
| BELM [40] | - | - | - | - | 0.6797 ±0.0038 | 0.7690 ±0.0417 | - |
| RVM [40] | - | - | - | - | 0.6783 ±0.0012 | 0.7750 ±0.0366 | - |
| PBL-McRBFN [41] | 0.7667 - | - | - | - | - | - | - |
| *Tanh-ELM* | 0.7602 ±0.0240 | 0.9507 ±0.0240 | 0.8142 ±0.0285 | 0.7855 ±0.0155 | 0.8675 ±0.0355 | 0.7642 ±0.0182 | 0.8960 ±0.0196 |
| *Rec-Tanh-ELM* | 0.7664 ±0.0249 | 0.9534 ±0.0173 | 0.8145 ±0.0264 | 0.7845 ±0.0158 | 0.8720 ±0.0369 | 0.7669 ±0.0186 | 0.8973 ±0.0191 |
| *ELM-BBFNN* | 0.7703 ±0.0238 | 0.9707 ±0.0183 | 0.8175 ±0.0294 | 0.7857 ±0.0157 | 0.8681 ±0.0317 | 0.7721 ±0.0147 | 0.9112 ±0.0183 |
| **Rec-ELM-BBFNN** | **0.7740** ±0.0201 | **0.9785** ±0.0170 | **0.8191** ±0.0241 | **0.7860** ±0.0156 | **0.8735** ±0.0347 | **0.7777** ±0.0155 | **0.9147** ±0.0153 |

accuracy under the best combination of parameters is shown and compared in table 4.

The improvement achieved by ELM-BBFNN over Tanh-ELM exceeds 14% for the earthquakes datasets. The upturn achieved by Rec-ELM-BBFNN over Rec-Tanh-ELM outstrips 12% for the same dataset. It is to be noticed that there is an accuracy betterment for our recurrent beta models over the feed-forward one. Table 5 presents the precision results realized not only by our approach but also by a number of other methods proposed in several previous researches applied to the breast cancer dataset.

According to the results listed within table 5, Rec-ELM-BBFNN performs better than the other models. The advance of our model is highly remarkable for this dataset. It overtakes many ELM- and SVM-based approaches. These results prove what have been already discussed and analyzed in the previous tables.

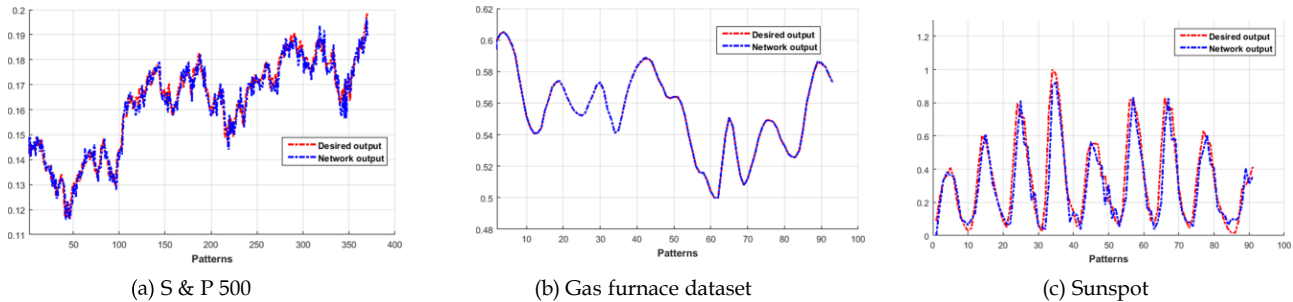

(a) S & P 500          (b) Gas furnace dataset          (c) Sunspot

Fig. 4. curves of the Rec-ELM-BBFNN's output and the target ones on a number of time series prediction datasets.

TABLE 3: Classification accuracy-based comparison with other existent approaches on the ECG200 dataset.

| Method | CA |
|---|---|
| STMF [42] | 0.700 |
| SVM [43] | 0.790 |
| LPP [44] | 0.710 |
| Swale [45] | 0.830 |
| SpADe [45] | 0.744 |
| GeTeM [45] | 0.800 |
| Gorecki's method [43] | 0.830 |
| FS [46] | 0.766 |
| ST+FCBF [46] | 0.766 |
| LPP [44] | 0.710 |
| NCC [42] | 0.770 |
| N5S2 [47] | 0.770 |
| EDTW [44] | 0.825 |
| *Tanh-ELM* | 0.744 ±0.036 |
| *Rec-Tanh-ELM* | 0.781 ±0.027 |
| *ELM-BBFNN* | 0.826 ±0.014 |
| *Rec-ELM-BBFNN* | **0.830** ±0.024 |

TABLE 4: Classification accuracy-based comparison with other existent approaches on the earthquakes dataset.

| Method | CA |
|---|---|
| ED [47] | 0.674 |
| DTWR [47] | 0.742 |
| N5S2 [47] | 0.807 |
| Euclidean1-NN [47] | 0.674 |
| MACD-SAX [47] | 0.792 |
| *Tanh-ELM* | 0.717 ±0.017 |
| *Rec-Tanh-ELM* | 0.728 ±0.016 |
| *ELM-BBFNN* | 0.817 ±0.021 |
| *Rec-ELM-BBFNN* | **0.819** ±0.010 |

## B. Time series prediction

Fig. 4 draws the curves of targeted signal and the output one and reveals the distinct proportionality between both of them on a number of time series datasets. It shows the proposed network's output signal and the desired one together. Based on fig. 4, it can be noticed that the predicted output curve fits the actual one well. Fig. 4 emphasizes the superposition between the output of our proposed network and the targets from the testing datasets. This conclusion is boosted by the numerical accuracy study performed in tables 6-9. According to what is included in these tables, Rec-ELM-BBFNN outperforms the other methods. This advance is traduced by attaining the least error in the majority of cases.

TABLE 5: Classification accuracy-based comparison with other existent approaches on the breast cancer dataset.

| Method | CA |
|---|---|
| 1-NN [36] | 0.957 |
| 3-NN [36] | 0.964 |
| C45 [36] | 0.948 |
| LDA [36] | 0.950 |
| CART with feature selection [48] | 0.946 |
| C.45 [49] | 0.948 |
| NB [50] | 0.857 |
| ESN [51] | 0.951 |
| ESN-anti-Oja [51] | 0.963 |
| ESN-BCM [51] | 0.978 |
| MLP1 [52] | 0.969 |
| CNN [52] | 0.978 |
| Ensemble FCLF-CNN [52] | 0.987 |
| Self-training [53] | 0.858 |
| ELM [37] | 0.962 ±0.0129 |
| ScELM [37] | 0.950 ±0.0060 |
| SVM [37] | 0.931 ±0.0184 |
| SBELM [40] | 0.972 ±0.0121 |
| TROP-ELM [40] | 0.970 ±0.0069 |
| BELM [40] | 0.970 ±0.0087 |
| RVM [40] | 0.973 ±0.0083 |
| SVM [40] | 0.973 ±0.0083 |
| SVM (gaussian kernel) [54] | 0.973 ±0.0083 |
| Unified ELM (gaussian kernel) [54] | 0.982 ±0.0083 |
| Sparse ELM (gaussian kernel) [54] | 0.982 ±0.0083 |
| *Tanh-ELM* | 0.986 ±0.0048 |
| *Rec-Tanh-ELM* | 0.984 ±0.0050 |
| *ELM-BBFNN* | 0.987 ±0.0034 |
| *Rec-ELM-BBFNN* | **0.994** ±0.0037 |

Tables 6-9 report the performances in terms of RMSE for the conceived approach (Rec-ELM-BBFNN) on a number of time series forecasting databases. In order to highlight the efficiency of our model, a comparison with other intelligent models is carried out in every table.

The RMSE-based comparison study performed in table 6 reveals the great performance of the Rec-ELM-BBFNN in dealing with a set of time series prediction tasks. The passenger dataset represents the international airline monthly total passengers' number in thousands from January 1949 to December 1960. The milk dataset includes the monthly milk production (pounds per cow) from 1962 to 1975. The stock problem encompasses the annual common stock price in US from 1871 till 1970.

TABLE 6: RMSE-based comparison with other existent approaches on a number of time series datasets.

| Method | Passenger | Milk | Stock | Lynx |
|---|---|---|---|---|
| SEANN [55] | 2.09 e+01 | - | - | - |
| TEANN [55] | 2.33 e+01 | - | - | - |
| ETS [56] | 5.24 e+01 | 2.67 e+01 | 6.97 e+00 | 4.72 e+01 |
| ARIMA [56] | 3.73 e+01 | 2.32 e+01 | 8.60 e+00 | 4.34 e+01 |
| MLP [57] | 3.34 e+01 | 1.42 e+01 | 3.23 e+01 | 1.09 e+03 |
| ARIMA-ANN [41] | 2.65 e+01 | 1.17 e+01 | 2.31 e+01 | 8.80 e+02 |
| ETSANN [57] | 2.17 e+01 | 6.41 e+00 | 7.18 e+00 | 8.82 e+02 |
| *Tanh-ELM* | 1.53 e+02 $\pm$ 1.22 e+01 | 4.52 e-01 $\pm$ 5.53 e-02 | 2.08 e+05 $\pm$ 4.77 e+03 | 1.26 e-01 $\pm$ 1.72 e-04 |
| *Rec-Tanh-ELM* | 8.91 e+01 $\pm$ 1.53 e+01 | 3.68 e-01 $\pm$ 1.85 e-01 | 2.01 e+01 $\pm$ 3.26 e+01 | 3.29 e-01 $\pm$ 1.67 e-01 |
| *ELM-BBFNN* | 4.06 e-01 $\pm$ 5.90 e-01 | 2.71 e-02 $\pm$ 1.74 e-02 | 1.01 e+00 $\pm$ 1.13 e+00 | 1.16 e-01 $\pm$ 2.40 e-03 |
| *Rec-ELM-BBFNN* | **1.82 e-01** $\pm$ 7.23 e-02 | **2.20 e-02** $\pm$ 8.50 e-03 | **3.16 e-01** $\pm$ 1.08 e-01 | **1.13 e-01** $\pm$ 1.30 e-03 |

The lynx database encloses the annual number of lynx trapped in MacKenzie River from 1821 to 1934.

For both of passenger and stock datasets, the advance of our beta-basis function model is highly noticeable as the error has dropped exponentially. Also, our recurrent network outperforms the feed-forward one as the RMSE decreases by 55%, 18% and 68% for passenger, stock and milk problems, respectively.

Table 7 includes the accuracy results of the implemented models in terms of RMSE on the S & P 500 U.S. financial dataset. This dataset permits to predict the future stock market price in U.S. These results are boosted by reporting the performances of other literature works that have focused on this problem.

TABLE 7: RMSE-based comparison with other existent approaches on the S & P 500 U.S. financial dataset.

| Method | RMSE |
|---|---|
| ESN [51] | 6.66 e-02 |
| ESN-anti-Oja [51] | 4.68 e-02 |
| ESN-BCM [51] | 4.34 e-02 |
| RSVM [58] | 2.45 e-01 |
| RELM [58] | 3.67 e-02 |
| RKERM [58] | 3.50 e-02 |
| RKELM+PSO [58] | 6.29 e-02 |
| RKERM+PSO [58] | 1.09 e-02 |
| Kalman Filter [59] | 1.00 e-02 |
| ARIMA [59] | 1.46 e-02 |
| ANN [59] | 9.74 e-02 |
| PSO+ANN [59] | 1.58 e-02 |
| TAEF [59] | 2.91 e-02 |
| *Tanh-ELM* | 8.30 e-03 $\pm$ 2.60 e-03 |
| *Rec-Tanh-ELM* | 1.21 e-02 $\pm$ 5.50 e-03 |
| *ELM-BBFNN* | 8.50 e-03 $\pm$ 3.00 e-03 |
| *Rec-ELM-BBFNN* | **8.10 e-03** $\pm$ 2.20 e-03 |

Based on table 7's content, the Rec-ELM-BBFNN outstrips significantly several approaches for the overwhelming majority of prediction and classification tasks. For this dataset, it realizes an amelioration rate over 25% compared to a set of already developed approaches in the literature. This same rate overtakes 2% compared to the Tanh-ELM, the Rec-Tanh-ELM and the ELM-BBFNN. As for the previous tests, the performances of several ELM variants are tracked in order to ensure the effectiveness of our proposed ELM-based model.

In order to go deeper in the empirical study, the testing RMSEs of the proposed approach and a set of intelligent existent approaches in the literature on the gas furnace problem are gathered in table 8.

In this dataset, the $CO_2$ concentration output $y(t)$ is predicted based on its previous value $y(t-1)$ and the gas flown $u(t-4)$.

TABLE 8: RMSE-based comparison with other existent approaches on the gas furnace dataset.

| Method | RMSE |
|---|---|
| HMDDE-BBFNN [16] | 2.41 e-01 |
| FBBFNT [17] | 1.81 e-01 |
| FWNN-M [60] | 2.32 e-02 |
| ODE [61] | 5.13 e-01 |
| WNN+ gradient [62] | 8.40 e-02 |
| WNN+ hybrid [62] | 8.10 e-02 |
| *Tanh-ELM* | 4.37 e-02 $\pm$ 2.50 e-03 |
| *Rec-Tanh-ELM* | 4.36 e-02 $\pm$ 2.30 e-03 |
| *ELM-BBFNN* | 4.36 e-02 $\pm$ 2.10 e-03 |
| *Rec-ELM-BBFNN* | **4.35 e-02** $\pm$ 1.90 e-03 |

Table 9 includes the two testing errors RMSE 1 and RMSE 2 achieved by a number of developed methods and ours on the sunspot number dataset. The test data are divided into two datasets: the first one is taken in between 1921 and 1955 while the second includes the remaining testing part. The output $y(t)$ is predicted based on previous four inputs which are $y(t-1)$, $y(t-2)$, $y(t-3)$ and $y(t-4)$. Table 9 ensures further the efficiency of the proposed method in terms of accuracy.

TABLE 9: RMSE-based comparison with other existent approaches on the sunspot number dataset.

| Method | RMSE 1 | RMSE 2 |
|---|---|---|
| ANFIS [63] | 1.91 e-01 | 4.06 e-01 |
| FWNN-S [64] | 3.30 e-01 | 5.20 e-01 |
| LFN [65] | 2.54 e-01 | 3.81 e-01 |
| H-MOEA RNNs [66] | 1.52 e-02 | - |
| H NARX-El RNN [67] | 1.19 e-02 | - |
| LFN [65] | 2.54 e-01 | 3.81 e-01 |
| LFN [65] | 2.54 e-01 | 3.81 e-01 |
| Kalman Filter [59] | - | 9.84 e-02 |
| ARIMA [59] | - | 1.47 e-01 |
| ANN [59] | - | 1.07 e-01 |
| PSO+ANN [59] | - | 9.77 e-02 |
| TAEF [59] | - | 1.08 e-01 |
| *Tanh-ELM* | 7.72 e-02 $\pm$ 6.80 e-03 | 1.19 e-01 $\pm$ 1.86 e-02 |
| *Rec-Tanh-ELM* | 7.81 e-02 $\pm$ 5.10 e-03 | 1.00 e-01 $\pm$ 1.98 e-02 |
| *ELM-BBFNN* | 7.69 e-02 $\pm$ 6.10 e-03 | 9.32 e-02 $\pm$ 1.73 e-02 |
| *Rec-ELM-BBFNN* | **7.46 e-02** $\pm$ 6.30 e-03 | **9.10 e-02** $\pm$ 1.30 e-02 |

## C. Regression

In order to analyze and emphasize the performance of the proposed network in large scale problems, a number of regression datasets are studied. These datasets are picked out from the KEEL archive [32]. Dealing with high dimensional data is a challenge for the proposed system to perform well and ensure its efficiency. The conceived network deals with three databases which are the ailerons, House and california.
A 5-fold cross validation method was performed for overall the tasks.
The ailerons dataset manages a flying problem of a F16 aircraft. It consists to forecast the control action on the aircraft's ailerons. The California problem contains information about all the block groups in California from the 1990 Census. The task is to approximate the median house value of each block from the values of the rest of the variables. The house dataset deals with the prediction of the median price of the house in a specific region. The attributes are related to the demographic composition as well as the state of housing market.
Table 10 presents a set of some methods which have worked on the ailerons, California and house datasets. Their results are compared to those achieved by "Rec-ELM-BFNN" approach. This last brings up the most reduced RMSE over the other methods.
For the house dataset, the accuracy improvement is considerably high. In fact, it is notable that the beta-based models have realized a big MSE drop. They perform with small standard deviations. The impact of the beta activation is revealed throughout this achievement. Also, the recurrent fashion has overtaken the feed-forward one. Indeed, Rec-ELM-BBFNN has realized a RMSE drop of more than 50% compared to the ELM-BBFNN model. To sum up, these results and others which have appeared in the last two subsections (4.1 and 4.2) answer the question 1 that has been already addressed. The efficacy of our studied approach is well revealed on a number of tasks.

TABLE 10: MSE-based comparison with other existent approaches on the ailerons and house datasets (Top:mean MSE and bottom: St.D).

| Method | Ailerons | House | California |
|---|---|---|---|
| $FSMOttFS^e + TUN^e$ [68] | - | 9.40 e+08 | 2.95 e+08 |
| | | | - |
| $MET SK - HD^\theta$ (first stage) [68] | - | 10.36 e+08 | 2.63 e+08 |
| | | | - |
| $MET SK - HD^\theta$ (final stage) [68] | - | 8.64 e+08 | 1.71 e+08 |
| | | | - |
| PAES-RL300 [69] | 2.47 e-08 | 1.05 e+09 | 2.95 e+09 |
| | ±8.11 e-09 | ±2.14 e+08 | ±2.27 e+08 |
| PAES-RL50 [69] | 4.52 e-08 | 1.20 e+09 | 3.20 e+09 |
| | ±2.38 e-08 | ±1.59 e+08 | ±2.80 e+08 |
| PAES-RCS [69] | 1.81 e-08 | 9.26 e+08 | 2.70 e+09 |
| | ±1.24 e-09 | ±1.05 e+08 | ±1.71 e+08 |
| PAES-RCS(10%) [69] | 1.90 e-08 | 9.39 e+08 | 2.70 e+09 |
| | ±1.70 e-09 | ±9.44 e+07 | ±1.33 e+08 |
| $V_M$ (3) [70] | 3.58 e-08 | - | - |
| | ±2.59 e-09 | | |
| $V_M$ (7) [70] | 4.67 e-08 | - | - |
| | ±1.98 e-09 | | |
| $FSMOttPS^e + TUN^e$ [70] | 2.00 e-08 | - | - |
| | ±2.74 e-09 | | |
| $FSMOttPS^e$ [70] | 2.36 e-08 | - | - |
| | ±2.36 e-09 | | |
| *Tanh-ELM* | 1.93 e-08 | 1.96 e+09 | 3.99 e+05 |
| | ±1.02 e-09 | ±6.69 e+09 | ±8.99 e+05 |
| *Rec-Tanh-ELM* | 1.90 e-08 | 1.18 e+09 | 1.32 e+05 |
| | ±1.38 e-09 | ±2.76 e+09 | ±2.92 e+05 |
| *ELM-BBFNN* | 1.83 e-08 | 3.74 e+04 | 5.80 e+04 |
| | ±1.97 e-09 | ±1.93 e+04 | ±1.97 e+03 |
| *Rec-ELM-BBFNN* | **1.75 e-08** | **3.35 e+04** | **5.74 e+04** |
| | ±1.34 e-09 | ±1.27 e+03 | ±1.24 e+03 |

### D. Robustness against noise

As it has been already mentioned, the robustness of the four implemented models is studied throughout supplying different Gaussian noise levels. Fig. 5 illustrates a statistical comparison study between Tanh-ELM, Rec-Tanh-ELM, ELM-BBFNN and Rec-ELM-BBFNN according to the noise levels squirted into a number of the already studied datasets.
Based on the results included in Fig.5, Rec-ELM-BBFNN shows bigger sturdiness against noise than the other implemented models. In fact, throughout the majority of noise levels, it persists in advance compared with the three implemented models. In fig. 5 (d), (e) and (f), the Rec-ELM-BBFNN's histograms indicate a higher CA while for the others (fig. 5 (a), (b), (c), (g), (h) and (i)), the Rec-ELM-BBFNN's histograms are the lowest ones. The histograms in fig. 5 (i) show a zoom from fig. 5 (h) of the results achieved by Rec-ELM-BBFNN and ELM-BBFNN on the passenger dataset. Hence, our proposed network, keeps a bigger CA for the classification tasks and a lower error for the time series prediction tasks in presence of noise. Thus, the added value of both of recurrent weights and Beta kernel is highly revealed in noise free as well as noisy data.
In order to beef up the empirical analysis, the statistical results already presented in Fig. 5 are boosted by a numerical analysis. In fact, the improvement rates $IR_1$, $IR_2$ and $IR_3$ given by our model Rec-ELM-BBFNN over the Tanh-ELM, Rec-Tanh-ELM and ELM-BBFNN, respectively are tracked

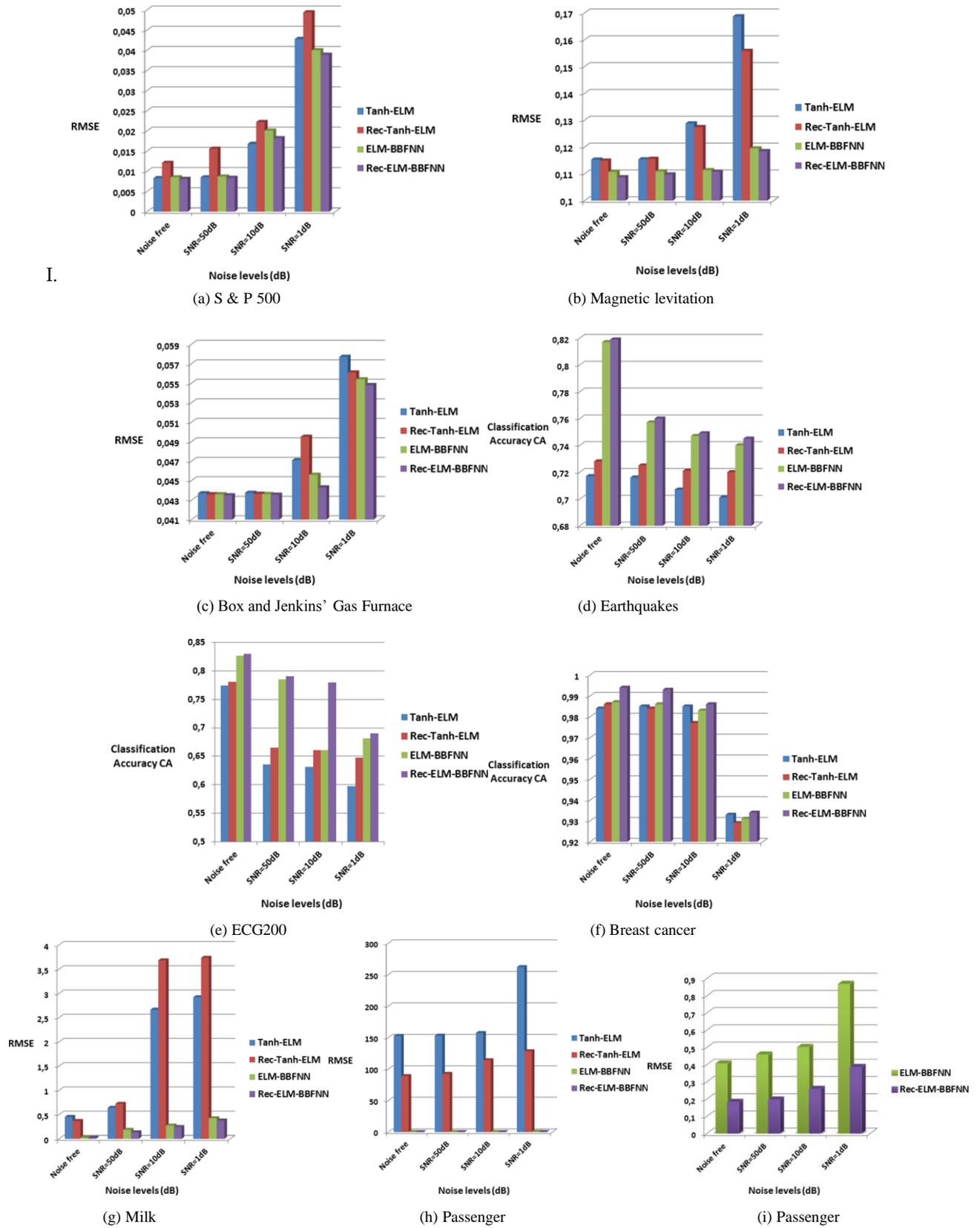

I.

(a) S & P 500

(b) Magnetic levitation

(c) Box and Jenkins' Gas Furnace

(d) Earthquakes

(e) ECG200

(f) Breast cancer

(g) Milk

(h) Passenger

(i) Passenger

Fig. 5. Statistical analysis of the performance evolution of Tanh-ELM, Rec-Tanh-ELM, ELM-BBFNN and Rec-ELM-BBFNN with and without addition of gaussian noise levels.

In the case of time series prediction or regression problems, the betterment incidence indicates the error (either RMSE or MSE) decrease, while for the classification tasks; the enhancement rate represents the increase of the classification accuracy (CA). For instance, in case of classification problems, the rates ($IR_1$, $IR_2$ and $IR_3$) are computed according to equations (17), (18) and (19). In the case of time series prediction or regression, just substitute the CA by RMSE or MSE to obtain these rates.

$$IR_1 = \frac{CA(Rec - ELM - BBFNN) - CA(Tanh - ELM)}{CA(Tanh - ELM)} \quad (17)$$

$$IR_2 = \frac{CA(Rec - ELM - BBFNN) - CA(Rec - Tanh - ELM)}{CA(Rec - Tanh - ELM)} \quad (18)$$

$$IR_3 = \frac{CA(Rec - ELM - BBFNN) - CA(ELM - BBFNN)}{CA(ELM - BBFNN)} \quad (19)$$

TABLE 11: Percentage of improvement rates $IR_1$, $IR_2$ and $IR_3$ realized by Rec-ELM-BBFNN against Tanh-ELM, Rec-Tanh- ELM and ELM-BBFNN, respectively (from top to down) on a number of datasets at different noise levels.

| Dataset | $IR_i$ | Noise=50dB | Noise=10dB | Noise=1dB |
|---|---|---|---|---|
| | $IR_1$ | 24.21 | 23.61 | 15.38 |
| ECG200 | $IR_2$ | 6.27 | 18.79 | 6.48 |
| | $IR_3$ | 0.48 | 0.63 | 1.32 |
| | $IR_1$ | 6.14 | 5.94 | 6.27 |
| Earthquakes | $IR_2$ | 4.82 | 3.88 | 3.47 |
| | $IR_3$ | 0.39 | 0.26 | 0.67 |
| | $IR_1$ | 0.81 | 0.10 | 0.10 |
| Breast cancer | $IR_2$ | 0.91 | 0.92 | 0.53 |
| | $IR_3$ | 0.70 | 0.30 | 0.32 |
| | $IR_1$ | 1.53 | 8.33 | 9.11 |
| S & P 500 | $IR_2$ | 57.05 | 18.01 | 21.25 |
| | $IR_3$ | 4.47 | 9.45 | 2.75 |
| | $IR_1$ | 4.85 | 13.97 | 29.75 |
| Magnetic lev. | $IR_2$ | 5.01 | 13.02 | 23.94 |
| | $IR_3$ | 0.09 | 0.53 | 0.83 |
| | $IR_1$ | 0.45 | 5.94 | 5.02 |
| Gas furnace | $IR_2$ | 0.22 | 10.50 | 2.31 |
| | $IR_3$ | 0.16 | 2.85 | 1.08 |
| | $IR_1$ | 99.87 | 99.83 | 99.70 |
| Passenger | $IR_2$ | 99.78 | 99.77 | 99.70 |
| | $IR_3$ | 57.21 | 48.63 | 55.62 |
| | $IR_1$ | 78.77 | 90.82 | 87.04 |
| Milk | $IR_2$ | 81.21 | 93.35 | 89.86 |
| | $IR_3$ | 27.82 | 10.89 | 10.58 |

Based on what report table 11 and fig. 5, the proposed recurrent beta basis model has ensured its overwhelmingly advance compared to the other implemented models. It has realized a remarkable learning strength, a noteworthy robustness and stability against exterior breakdowns. Thus, the empirical study given in this subsection represents an answer to the question 2.

Throughout the experimental study, the proposed recurrent BBFNN shows an outstanding capability to achieve high accuracy levels for the studied tasks. Overall, in the majority of tasks (classification, prediction and regression), it has outperformed many already designed models in the literature. The power of the beta function as a kernel of neural networks is obviously revealed. Also, the added value of recurrent connections seems to be very important as it has performed better than the feed-forward models for overall the datasets.

## V. Conclusion

Throughout this paper, the BBFNN is thoroughly studied. Both of the training and architecture of BBFNN are altered to give birth to the Rec-ELM-BBFNN. This last presents a new recurrent variant of BBFNN where ELM has been used as a training method and the recurrent fashion characterized its architecture.

Previously, the inputs of the BBFNN are transformed as they are to the beta feature space. Thus, all the inputs have the same impact in the computation of the network output. With the use of ELM, the contribution of the input is mainly related to a random input weight value the fact that gives every input a distinguished impact on the hidden layer.

To boost the architecture of the BBFNN, a new recurrent inner weight matrix is defined in between the neurons of the beta layer. This matrix addition creates a kind of virtual memory about the past hidden activations. Hence, more information are provided to the network which becomes more dynamic and capable to deal with more nonlinear complex tasks than before. The proposed Rec-ELM-BBFNN is tested on a set of tasks including classification, regression and time series prediction.

According to the empirical study, the new recurrent beta network realizes considerable advance compared to the feed-forward beta network. Beta kernel function overtakes, for the majority of datasets, the well-known hyperbolic tangent transfer function. The proposed model has ensured a great sturdiness against the breakdowns squirted in a number of datasets. As for future work, many other ELM variants are under investigation to be used to train BBFNN.

## Acknowledgment

The research leading to these results has received fund- ing from the Ministry of Higher Education and Scientific Research of Tunisia under the grant agreement number LR11ES48.